\documentclass[10pt,twocolumn,journal]{IEEEtran}
\usepackage{xcolor}
\usepackage{float}
\usepackage{mathrsfs}
\usepackage{amsfonts}
\usepackage{amssymb}
\usepackage{amsthm}
\usepackage{epsfig}
\usepackage{graphicx}
\usepackage{graphics}
\usepackage{bm}
\usepackage{amsmath,latexsym,amsthm,array,algorithm,algpseudocode,booktabs}
\usepackage{cite}
\usepackage{subfig}
\usepackage{multirow}
\usepackage{makecell}
\usepackage{subfig}
\usepackage{longtable}
\usepackage{threeparttable}

\ifCLASSINFOpdf
\else
\fi
\begin{document}
%
\title{\textbf{Multi-task GANs for Semantic Segmentation and Depth Completion with Cycle Consistency}}
%
%
%
\author{Chongzhen Zhang, 
	    Yang Tang, \textit{Senior Member, IEEE,}\\
	    Chaoqiang Zhao,
	    Qiyu Sun,
        Zhencheng Ye
        and J\"{u}rgen Kurths
        
\thanks{This work was supported by the National Key Research and Development Program of China under Grant 2018YFC0809302, the National Natural Science Foundation of China under Grant Nos. 61988101, 61751305, 61673176, by the Programme of Introducing Talents of Discipline to Universities (the 111 Project) under Grant B17017. (\textit{Corresponding authors: Yang Tang, Zhencheng Ye.})}
\thanks{C. Zhang,  Y. Tang, C. Zhao, Q. Sun and Z. Ye are with the Key Laboratory of Advanced Control and Optimization for Chemical Process, Ministry of Education, East China University of Science and Technology, Shanghai 200237, China (e-mail: yangtang@ecust.edu.cn (Y. Tang) and yzc@ecust.edu.cn (Z. Ye)).}
\thanks{J. Kurths is with the Potsdam Institute for Climate Impact Research, 14473 Potsdam, Germany, and with the Institute of Physics, Humboldt University of Berlin, 12489 Berlin, Germany (e-mail: juergen.kurths@pik-potsdam.de).}

}

\markboth{Multi-task GANs for Semantic Segmentation and Depth Completion with Cycle Consistency} {Shell
\MakeLowercase{\textit{et al.}}: Bare Demo of IEEEtran.cls for
Journals}
\maketitle

\begin{abstract}
Semantic segmentation and depth completion are two challenging tasks in scene understanding, and they are widely used in robotics and autonomous driving. Although several studies have been proposed to jointly train these two tasks using some small modifications, like changing the last layer, the result of one task is not utilized to improve the performance of the other one despite that there are some similarities between these two tasks. In this paper, we propose multi-task generative adversarial networks (Multi-task GANs), which are not only competent in semantic segmentation and depth completion, but also improve the accuracy of depth completion through generated semantic images. In addition, we improve the details of generated semantic images based on CycleGAN by introducing multi-scale spatial pooling blocks and the structural similarity reconstruction loss. Furthermore, considering the inner consistency between semantic and geometric structures, we develop a semantic-guided smoothness loss to improve depth completion results. Extensive experiments on Cityscapes dataset and KITTI depth completion benchmark show that the Multi-task GANs are capable of achieving competitive performance for both semantic segmentation and depth completion tasks.

\end{abstract}


%
\IEEEpeerreviewmaketitle

\section{Introduction}
	
In the past decades, the computer vision community has implemented a large number of applications in autonomous systems \cite{shao2014transfer}. As a core task of autonomous robots, scene perception involves several tasks, including semantic segmentation, depth completion, and depth estimation, etc \cite{zhang2020autonomous}. Due to the great performance of deep learning (DL) in image processing, several methods based on convolutional neural networks (CNNs) have achieved competitive results in perception tasks of autonomous systems, like semantic segmentation \cite{liu2018collaborative}, object detection \cite{shih2019real}, image classification \cite{zhu2020deep} and depth completion \cite{zhao2020monocular}, \cite{tang2020overview}. DL-based visual perception methods are usually divided into supervised, semi-supervised and unsupervised ones according to the supervised manner. Since the ground truth is difficult to acquire, perceptual tasks by semi-supervised and unsupervised methods have attracted increasing attention in recent years. Each pixel in one image usually contains rich semantic and depth information, which makes it possible to implement semantic or depth visual perception tasks based on the same framework \cite{zhang2020autonomous}. Since most unsupervised depth estimation methods predict depth information from stereo image pairs or image sequences, which only focus on RGB image information, and they may suffer from a similar over-fitting outcome \cite{zhao2020monocular}. In contrast, depth completion tasks sparse depth as references, which is capable of achieving better performance than depth estimation from RGB \cite{zou2020simultaneous}. Therefore, we consider depth completion instead of depth estimation in this paper. We propose a multi-task framework, including semantic segmentation and depth completion. 

\begin{figure}[t]
	\centering
	\includegraphics[width=9cm,height=5cm]{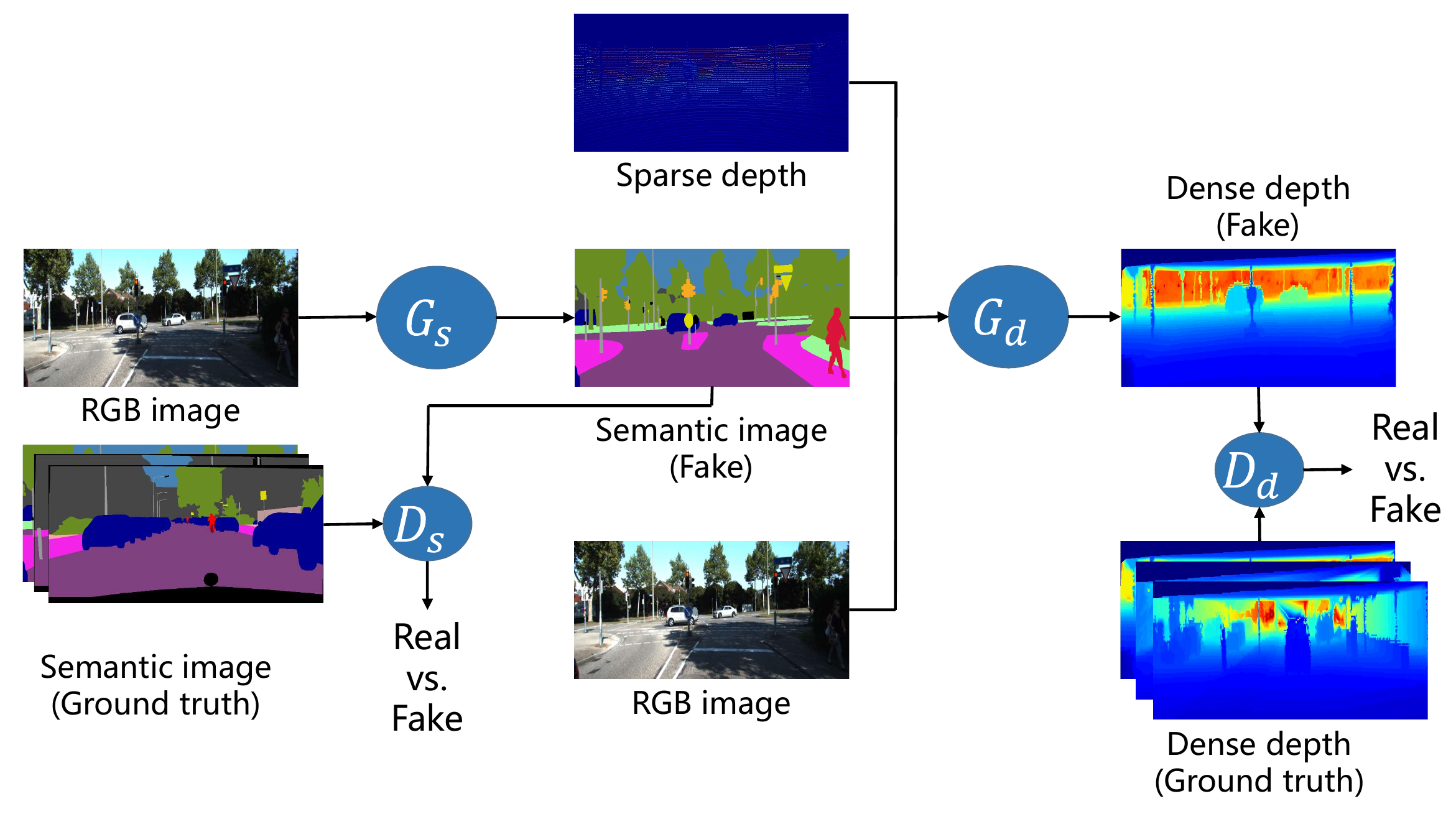}
	\caption{The proposed multi-task generative adversarial networks. Our multi-task framework includes semantic preception task and depth completion task, and each task corresponds to its own generator ($G$) and discriminator ($D$). We use the results of semantic segmentation to improve the accuracy of depth completion.}
	\label{Fig1}
\end{figure}

As an important way for autonomous systems to understand the scene, semantic segmentation is a pixel-level prediction task, which classifies each pixel into a corresponding label \cite{garcia2017review}. The accuracy of supervised semantic segmentation methods is satisfactory, while ground truth labels are usually difficult to obtain. Therefore, several semantic segmentation algorithms have focused on unsupervised methods in recent years, like FCAN \cite{zhang2018fully}, CyCADA \cite{hoffman2018cycada} and CrDoCo \cite{chen2019crdoco}. Unsupervised semantic segmentation methods use semantic labels of simulated datasets without manual annotation, and they eliminate the domain shift between real-world data and simulated data through domain adaptation. In other words, although unsupervised methods do not require manual labeling, they still constrain pixel classification in a supervised manner by automatically generating labels paired with RGB images \cite{hoffman2018cycada}, \cite{pan2020unsupervised}. In this paper, we treat semantic segmentation tasks that do not require paired semantic labels, i.e., we regard semantic segmentation as a semi-supervised image-to-image translation task \cite{zhu2017unpaired}. Actually, this is a challenging task since we utilize unpaired RGB images and semantic labels in the training process may permute the labels for vegetation and building \cite{zhu2017unpaired}. Our generated semantic images may not be strictly aligned with the category labels, but we align generated semantic images with category labels by minimizing the pixel value distance \cite{zhu2017unpaired}. 


Image-to-image translation is to transfer images from the source domain to the target domain. In this process, we should ensure that the content of translated images is consistent with the source domain, and the style is consistent with the target domain \cite{zhang2020autonomous}. In image-to-image translation, supervised methods are able to produce good transfer results. However, it is very difficult to obtain paired data of different styles \cite{zhu2017unpaired}. Therefore, image translation using unpaired data has attracted much attention in recent years. Since generative adversarial networks (GANs) have shown powerful effects in image generation tasks, Zhu \textit{et al.}, \cite{zhu2017unpaired} present a cycle-consistent adversarial network (CycleGAN), which introduces the adversarial idea for image-to-image translation. CycleGAN is also used for semantic segmentation, while it has a poor effect on image details, like buildings and vegetation are often turned over due to the use of unpaired data \cite{zhu2017unpaired}. Li \textit{et al.} \cite{li2019asymmetric} improve the details of the generated images by adding an encoder and a discriminator into the architecture to learn an auxiliary variable, which effectively tackles the permutation problem of vegetation and buildings. Although the method in \cite{li2019asymmetric} is competitive, the introduction of the additional encoder and discriminator greatly increases the complexity of the model.  In order to improve the details of the generated images and deal with the permutation problem of vegetation and buildings, we introduce here multi-scale spatial pooling blocks \cite{tang2019multi} and the structural similarity reconstruction loss \cite{wang2004image}, which can achieve competitive results by adding minor complexity of the model only.  


Dense and accurate depth information is essential for autonomous systems, including tasks like obstacle avoidance \cite{zhang2014loam} and localization \cite{wolcott2015fast}, \cite{ma2019self}. With its high accuracy and perception range, LiDAR has been integrated into a large number of autonomous systems. Existing LiDAR provides sparse measurement results, and hence depth completion, which estimates the dense depth from sparse depth measurements, is very important for both academia and industry \cite{ma2019self}, \cite{qiu2019deeplidar}. Since the sparse depth measured by LiDAR is irregular in space \cite{ma2019self}, the depth completion is a challenging task. Existing multiple-input methods predict a dense depth image from the sparse depth and corresponding RGB image, while the RGB image is very sensitive to optical changes, which may affect the results of depth completion  \cite{chen2019towards}. In this paper, we are making full use of semantic information to implement depth completion tasks, which substantially reduces the sensitivity to optical changes \cite{chen2019towards}.

In this paper, considering the fact that most studies require corresponding semantic labels as pixel-level constraints for semantic segmentation tasks \cite{zhang2018fully}, \cite{hoffman2018cycada}, \cite{chen2019crdoco}, we develop image translation methods to implement semantic segmentation with unpaired datasets. As well, we tackle the problem that vegetation and buildings are permuted in generated semantic images, which is widely observed in previous works \cite{zhu2017unpaired}, \cite{fu2019geometry}, by introducing multi-scale spatial pooling blocks \cite{tang2019multi} and the structural similarity reconstruction loss \cite{wang2004image}. In addition, although semantic cues are important for depth completion \cite{lu2020depth}, existing methods do not consider input semantic labels. We further introduce semantic information to improve the accuracy of depth completion and aim to achieve competitive results. Finally, we unify the two tasks in one framework. Our architecture diagram is shown in Figure \ref{Fig1}.



In summary, our main contributions are: 
\begin{itemize}
	\item For semantic segmentation, we introduce multi-scale spatial pool blocks to extract image features from different scales to improve the details of generated images. As well, we develop the structural similarity reconstruction loss, which improves the quality of reconstructed images from the perspectives of luminance, contrast and structure. 
	\item We consider the information at the semantic-level to effectively improve the accuracy of depth completion, by extracting the features of generated semantic images as well as by regarding the semantic-guided smoothness loss, which reduces the sensitivity to optical changes due to the consistency of object semantic information under different lighting conditions.
	\item We propose Multi-task GANs, which unify semantic segmentation and depth completion tasks into one framework, and experiments show that, compared with the state-of-the-art methods, Multi-task GANs achieve competitive results for both tasks, including CycleGAN \cite{zhu2017unpaired} and GcGAN \cite{fu2019geometry} for semantic segmentation, Sparse-to-Dense(gd) \cite{ma2019self} and NConv-CNN \cite{eldesokey2019confidence} for depth completion. 

\end{itemize}

The organization of this paper is arranged as follows. Section \ref{sec.2} discusses previous works on semantic segmentation, depth completion, and image-to-image translation. The method of this paper is introduced in Section \ref{sec.3}. Section \ref{sec.4} shows the experimental results of our proposed method on the Cityscapes dataset and KITTI depth completion benchmark. Finally, Section \ref{sec.5} concludes this study.

\section{Related work}\label{sec.2}

\textbf{Image-to-image translation.} The image-to-image translation task is to transform an image from the source domain to the target domain. It ensures that the content of the translated image is consistent with the source domain and the style is consistent with the target domain \cite{gatys2016image}. GANs use adversarial ideas for generating tasks, which can output realistic images \cite{goodfellow2014generative}. Zhu \textit{et al.} \cite{zhu2017unpaired} propose CycleGAN for the image-to-image translation task with unpaired images, which combines GANs and cycle consistency loss. In this paper, we consider the problem of semantic segmentation by using image-to-image translation. CycleGAN looks poor in the details of generating semantic labels from RGB images, e.g., buildings and vegetation are sometimes turned over \cite{zhu2017unpaired}. Fu \textit{et al.} \cite{fu2019geometry} develop a geometry-consistent generative adversarial network (GcGAN), which takes the original image and geometrically transformed image as input to generate two images with the corresponding geometry-consistent constraint. Although GcGAN improves the accuracy of generated semantic images, it does not tackle the permutation problem of buildings and vegetation efficiently. In order to deal with the problem of label confusion and further improve accuracy, Li \textit{et al.} \cite{li2019asymmetric} present an asymmetric GAN (AsymGAN), which improves the details of the generated semantic images. Li \textit{et al.} introduce an additional encoder and discriminator to learn an auxiliary variable, which greatly increases the complexity of the model. In order to improve the details of generated images and handle the problem that labels of vegetation and trees are sometimes turned over, we introduce multi-scale spatial pooling blocks and a structural similarity reconstruction loss without adding additional discriminators or encoders. Because multi-scale spatial pooling blocks extract features of different scales, and the structural similarity reconstruction loss improves the quality of generated images from the perspectives of luminance, contrast and structure.

\textbf{Semantic segmentation.} As a pixel-level classification task, semantic segmentation generally requires ground truth labels to constrain segmentation results through cross-entropy loss. Long \textit{et al.}  \cite{long2015fully} is the first to use fully convolutional networks (FCNs) for semantic segmentation tasks, and they add a skip architecture to improve the semantic and spatial accuracy of the output. This work is regarded as a milestone for semantic segmentation by DL \cite{zhang2020autonomous}. Inspired by FCN, several frameworks have been proposed to improve the performance of FCN, like RefineNet \cite{lin2017refinenet}, DeepLab \cite{chen2017deeplab}, PSPNet \cite{zhao2017pyramid}, etc. Since manual labeling is costly, several studies have focused on unsupervised semantic segmentation recently \cite{hoffman2018cycada}, \cite{chen2019crdoco}, \cite{hoffman2016fcns}. Unsupervised semantic segmentation usually requires synthetic datasets, like Grand Thief Auto (GTA) \cite{richter2016playing}, which can automatically label semantic tags of pixels without manual labor. However, due to the domain gap between synthetic datasets and real-world data, unsupervised methods should adapt to different domains \cite{hoffman2018cycada}, \cite{hoffman2016fcns}. Zhang \textit{et al.} \cite{zhang2018fully} introduce a fully convolutional adaptive network (FCAN), which combines appearance adaptation networks and representation adaptation networks for semantic segmentation. Due to the powerful effect of GANs in image generation and style transfer, Hong \textit{et al.} \cite{hong2018conditional} consider using cGAN \cite{mirza2014conditional} for domain adaptation in semantic segmentation. 
Aligning these two domains globally through adversarial learning may cause some categories to be incorrectly mapped, so Luo \textit{et al.} \cite{luo2019taking} introduce a category-level adversarial network to enhance local semantic information. The above semantic segmentation methods all need corresponding semantic tags as supervision signals for accurate classification, which requires paired RGB images and semantic image datasets. 
In this paper, we consider image-to-image translation for semantic segmentation, which does not require pairs of RGB images and semantic labels. This is a challenging task because the use of unpaired datasets may cause confusion in semantic labels \cite{zhu2017unpaired}.

\textbf{Depth completion.} Depth completion is to predict a pixel-level dense depth from the given sparse depth. Existing depth completion algorithms are mainly divided into depth-only methods and multiple-input methods \cite{lu2020depth}. Depth-only methods may provide the corresponding dense depth image by only inputting the sparse depth, which is a challenging problem due to the lack of rich semantic information. Uhrig \textit{et al.} \cite{uhrig2017sparsity} consider a sparse convolution module that operates on sparse inputs, which uses a binary mask to indicate whether the depth value is available. However, this method has a limited improvement of performance in deeper layers. Lu \textit{et al.} \cite{lu2020depth} show recovering some image semantic information from sparse depth. The depth completion model presented by Lu \textit{et al.} takes sparse depth as the only input, and it constructs a dense depth image and a reconstructed image simultaneously. This method can overcome the shortcomings that depth-only methods may fail to recover semantically consistent boundaries. Unlike depth-only methods, multiple-input methods generally take sparse depth images and corresponding RGB images as input, such that the model can make appropiate use of the rich semantic and geometric structure information in images. Ma \textit{et al.} \cite{mal2018sparse} consider inputting the concatenation of the sparse depth and the corresponding image to a deep regression model to obtain dense depth. Eldesokey \textit{et al.} \cite{eldesokey2019confidence} improve the normalized convolutional layer for CNNs with a highly sparse input, which treats the validity mask as a continuous confidence field. This method implements depth completion with a small number of parameters. In addition, several methods use other information to enhance the depth completion results, like surface normals \cite{qiu2019deeplidar} and semantic information \cite{jaritz2018sparse}.
Most existing studies consider multiple-input to extract richer image information, they do not include the sensitivity of RGB images to optical changes \cite{qiu2019deeplidar}, \cite{chen2019towards}, \cite{mal2018sparse}. In this paper, we consider the sensitivity of RGB images to optical changes, in which we add semantic-level feature extraction and smoothness loss.

\textbf{Multi-task learning.} Multi-task learning improves the performance of individual learning by incorporating  different but related tasks and sharing features between different tasks \cite{lu2020depth}, \cite{argyriou2007multi}. Especially in the field of image processing, each image contains rich geometric and semantic information. Therefore, multi-task learning has been applied to computer vision tasks, like semantic segmentation \cite{pham2019jsis3d} and depth estimation \cite{zhang2019pattern}. Qiu \textit{et al.} \cite{qiu2019deeplidar} introduce a framework for depth completion from sparse depth and RGB images, which uses the surface normal as the intermediate representation. They combine estimated surface normals and RGB images with the learned attention maps to improve the accuracy of depth estimation. Jaritz \textit{et al.} \cite{jaritz2018sparse} jointly train the network for depth completion and semantic segmentation by changing only the last layer, which is a classic multi-task method for semantic segmentation and depth completion. However, Jaritz \textit{et al.} do not further consider improving accuracy between tasks, i.e., they do not use semantic results to improve the accuracy of depth completion, or use depth completion results to improve semantic segmentation accuracy. Since semantic cues are important for depth completion \cite{lu2020depth}, we unify semantic segmentation and depth completion into a common framework, and we use the results of semantic segmentation to improve the accuracy of depth completion. As mentioned in \cite{zhang2017survey}, multi-task learning is challenging to set the network architecture and loss function. The network should consider all tasks at the same time, and cannot let simple tasks dominate the entire framework. Moreover, the weight distribution of the loss function should make all tasks equally important. If the above two aspects cannot be taken into consideration at the same time, the network may not converge or the performance may deteriorate. Meeting the above two requirements at the same time is challenging for multi-task. In this paper, we propose Multi-task GANs, which do not share the network layer for semantic segmentation and depth completion. Nevertheless, it still has coupling relationships between tasks, hance we regard Multi-task GANs as a multi-task framework.

Motivated by the above discussions, our method is not only competent in semantic segmentation and depth completion, but also effectively uses the semantic information to improve the accuracy of depth completion. For semantic segmentation, we regard it as an image-to-image translation task, and we improve the poor details of generated images, like pedestrians, vegetation and building details. For depth completion, we use the generated semantic images to improve its accuracy from the perspectives of feature extraction and loss function. The experiments in this paper verify the effectiveness of the propsoed framework both qualitatively and quantitatively.

\section{Methodology}\label{sec.3} 

In this section, we will first introduce our multi-task framework, including semantic segmentation module and depth completion module. Then we will introduce the loss function of each task.

\begin{figure*}[htbp]
	\centering
	\includegraphics[width=12.5cm,height=6.9cm]{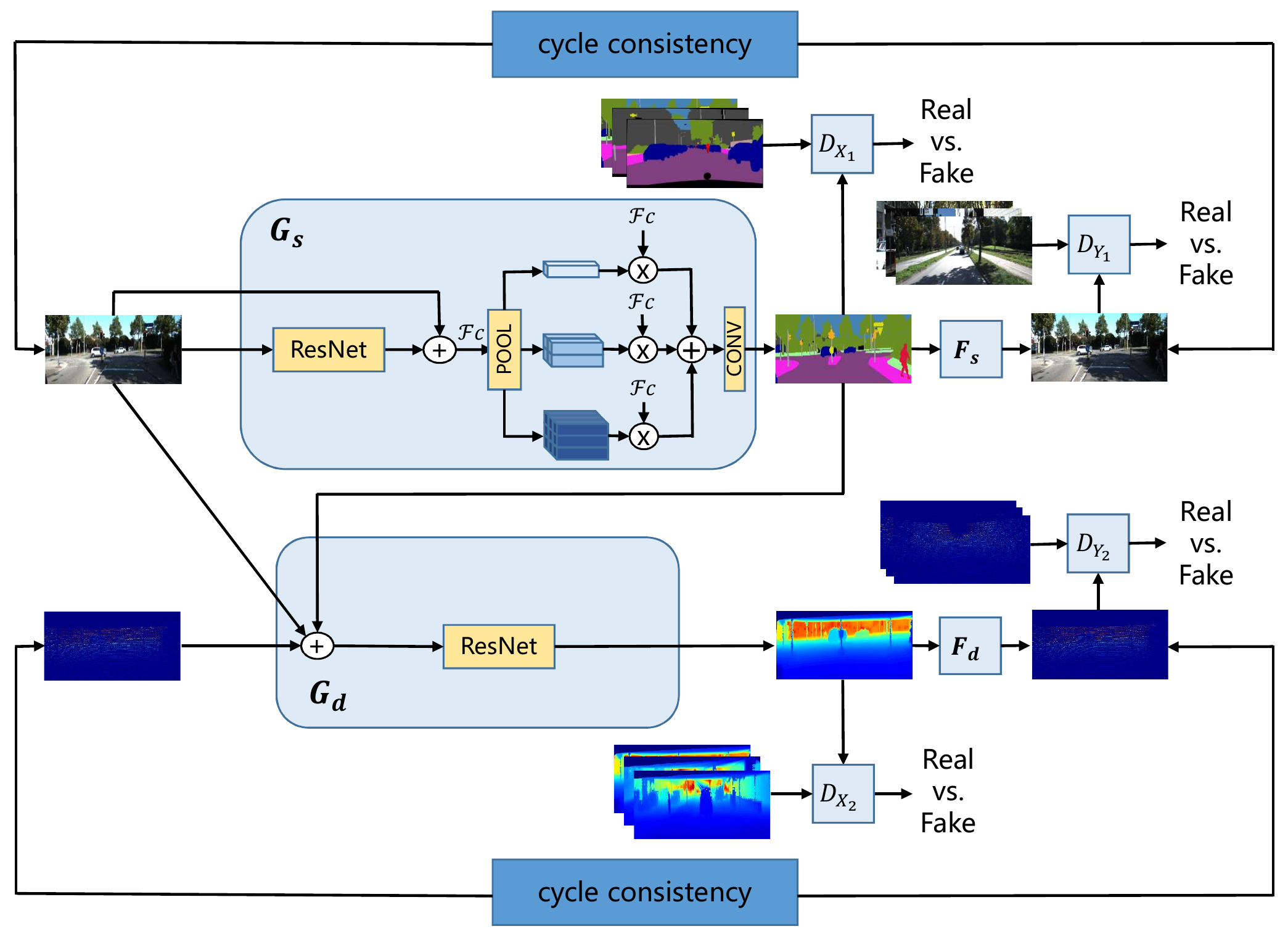}
	\caption{Architecture overview. Our multi-task framework contains two branches: semantic segmentation and depth completion.  The semantic segmentation branch contains the generators $G_s$, $F_s$ and the discriminators  $D_{X_{\scriptsize 1}}$, $D_{Y_{\scriptsize 1}}$. The depth completion branch contains the generators $G_d$, $F_d$ and the discriminators  $D_{X_{\scriptsize 2}}$, $D_{Y_{\scriptsize 2}}$.  We briefly show the network architectures of $F_s$ and $F_d$ in this figure, which have the same architectures as $G_s$ and $G_d$ respectively. The symbols $\bigoplus$ and $\bigotimes$ denote element-wise multiplication and concatenation, respectively.}
	\label{Fig2}
\end{figure*}

\subsection{Architecture overview}

Our multi-task framework is shown in Figure \ref{Fig2}. Our multi-task GANs effectively unify semantic segmentation and depth completion, and utilize semantic information to improve the depth completion results. We introduce multi-scale pooling blocks to extract features of different scales to improve the details of generated images. The architecture includes two branches: semantic segmentation and depth completion. The semantic segmentation branch translates the RGB image into a semantic label through the generator $G_s$, and then reconstructs it back to the RGB image through $F_s$. The discriminators $D_{X_{\scriptsize 1}}$ and $D_{Y_{\scriptsize 1}}$ are used to discriminate the generated semantic image and the reconstructed RGB image, respectively. Similarly, the depth completion branch translates the concatenation of sparse depth, RGB image and generated semantic image into a dense depth image through the generator $G_d$, and then reconstructs it back to the sparse depth through $F_d$. The discriminators $D_{X_{\scriptsize 2}}$ and $D_{Y_{\scriptsize 2}}$ are used to discriminate the generated dense depth and the reconstructed sparse depth, respectively. It is worth noting that our framework unifies semantic segmentation and depth completion tasks, and we introduce semantic information to improve the results of depth completion. Our generators for such tasks have only a small modification, i.e., adding a multi-scale pooling block for semantic segmentation based on the architecture of depth completion generator. Moreover, our discriminators for the two tasks are exactly the same.

\subsection{Semantic segmentation}

Our semantic segmentation task categorizes image pixels in the way of image-to-image translation. Since it is not strictly classified into the color value corresponding to the label, we process the translation result to align it to the color value of the label by calculating the pixel value distance from the standard semantic labels \cite{zhu2017unpaired}.

\textbf{Network.} We modify the architecture for our semantic segmentation generators from Zhu \textit{et al.} \cite{zhu2017unpaired}, which shows impressive results in image style transfer. Our semantic branch implements the translation between the RGB image domain $X_1$ and the semantic label domain $Y_1$ with unpaired training examples ${x} \in X_1$ and ${y_s} \in Y_1$. Our semantic segmentation module contains two generators $G_{s}: X_1 \rightarrow Y_1$ and  $F_{s}: Y_1 \rightarrow X_1$. We introduce multi-scale spatial pooling blocks \cite{tang2019multi} in generators, which capture sufficient spatial information when there is a large scene deformation between the source domain and the target domain \cite{tang2019multi}. Instead of cascading the last convolution layers of the generator into the multi-scale spatial pooling block, we concatenate the output of the residual blocks with the input RGB image, as shown in Figure \ref{Fig2}.  The structure of $F_s$ is the same as $G_s$. In addition, the discriminators $D_{X_{\scriptsize 1}}$ and $D_{Y_{\scriptsize 1}}$ are used to distinguish real samples from $x$ and $y_s$ or generated samples $F(y_s)$ and $G(x)$. Letting $\mathcal{L}_{seg} \triangleq \mathcal{L}_{seg}(G_s, F_s, D_{X_{\scriptsize 1}}, D_{Y_{\scriptsize 1}})$, the formula of the overall objective loss function of our semantic segmentation branch is then as follows:

\begin{equation}\label{seg_all_loss}
	\begin{aligned}
	\mathcal{L}_{seg} = & \mathcal{L}_{GAN}(G_s, D_{Y_{\scriptsize 1}}) + \mathcal{L}_{GAN}(F_s, D_{X_{\scriptsize 1}}) \\ & + \lambda_1 \mathcal{L}_{cyc}(G_s, F_s) + \lambda_2 \mathcal{L}_{rec}(G_s, F_s), 
	\end{aligned}
\end{equation}
where $\mathcal{L}_{GAN}(G_s, D_{Y_{\scriptsize 1}})$ and $\mathcal{L}_{GAN}(F_s, D_{X_{\scriptsize 1}})$ denote adversarial losses \cite{goodfellow2014generative}. $\mathcal{L}_{cyc}(G_s, F_s)$ refers to the cycle consistency loss proposed by Zhu \textit{et al.} \cite{zhu2017unpaired}. $\mathcal{L}_{rec}(G_s, F_s)$ is the structural similarity reconstruction loss, which is inspired by Wang \textit{et al.} \cite{wang2004image}. $\lambda_1$ and $\lambda_2$ are hyperparameters used to control the relative importance of cycle consistency loss and structural similarity reconstruction loss \cite{wang2004image}. 
We aim to solve:
\begin{equation}\label{seg_aim}
\begin{aligned}
G_s^*, F_s^* = \text{arg} \min\limits_{G_s, F_s} \max\limits_{D_{X_{\scriptsize 1}}, D_{Y_{\scriptsize 1}}}{L}_{seg}(G_s, F_s, D_{X_{\scriptsize 1}}, D_{Y_{\scriptsize 1}}).
\end{aligned}
\end{equation}

\textbf{Adversarial loss.} We employ adversarial losses \cite{goodfellow2014generative} for the semantic segmentation task. For the generator $G_{s}: X_1 \rightarrow Y_1$ and its corresponding discriminator $D_{Y_{\scriptsize 1}}$, the objective is formulated as follows:
\begin{equation}\label{GAN_loss}
\begin{aligned}
\mathcal{L}_{GAN}(G_{s},D_{Y_{\scriptsize 1}}) = &\mathbb{E}_{y_s \sim p_{\scriptsize \rm data}(y_s)}[\log D_{Y_{\scriptsize 1}}(y_s)]\\ &+ \mathbb{E}_{x \sim p_{\scriptsize \rm data}(x)}[\log (1-D_{Y_{\scriptsize 1}}(G_{s}(x))].
\end{aligned}
\end{equation}
We apply a similar adversarial loss $\mathcal{L}_{GAN}(F_s, D_{X_{\scriptsize 1}})$ for $F_{s}: Y_1 \rightarrow X_1$ and its corresponding discriminator $D_{X_{\scriptsize 1}}$.

\textbf{Cycle consistency loss.} We apply the cycle consistency loss \cite{zhu2017unpaired} to ensure the consistency between the RGB image domain and the semantic domain. In other words, we make that for each RGB image, after passing through the generators $G_{s}$ and $F_s$, the resulting image is as consistent as possible with the input image, i.e., $x \rightarrow G_s(x) \rightarrow F_s(G_s(x)) \approx x$. Similarly, we make that $y_s \rightarrow F_s(y_s) \rightarrow G_s(F_s(y_s)) \approx y_s$. The cycle consistency loss is formulated as:
\begin{equation}\label{cycleGAN_loss}
\begin{aligned}
\mathcal{L}_{cyc}(G_s, F_s) = &\mathbb{E}_{x \sim p_{\scriptsize \rm data}(x)}[\left\|F_s(G_s(x))-x\right\|_1]\\ &+ \mathbb{E}_{y_s \sim p_{\scriptsize \rm data}(y_s)}[\left\|G_s(F_s(y_s))-y_s\right\|_1].
\end{aligned}
\end{equation}

\textbf{Structural similarity reconstruction loss.} To further improve the accuracy of image-to-image translation, we regard $ \hat{x} = F_s(G_s(x))$ as the reconstructed image of $x$ generated by $G_{s}$ and $F_{s}$, and we use SSIM \cite{wang2004image} to constrain the luminance, contrast and structure of $\hat{x}$. Similarly, we regard $ \hat{y}_s = G_s(F_s(y_s))$ as the reconstructed image of the input $y_s$ generated by the $F_{s}$ and $G_{s}$, and we also use the reconstruction loss to constrain it. The reconstruction loss is formulated as:
\begin{equation}\label{ssim}
\begin{aligned}
\mathcal{L}_{rec}(G_s, F_s) = (1-SSIM(x, \hat{x})) + (1-SSIM(y_s, \hat{y}_s)).
\end{aligned}
\end{equation}

\subsection{Depth completion}

Generally, depth completion algorithms are mainly classified into depth-only methods and multiple-input methods, where multiple-input methods often take RGB images and sparse depth as inputs. Different from the previous works, like \cite{ma2019self} and \cite{qiu2019deeplidar}, we take the generated semantic image $\widetilde{y}_s$ from the semantic branch into the depth completion module together with RGB image and sparse depth image.

\textbf{Network.} 
We concatenate the sparse depth $x_d$, color image $x_i$, and generated semantic image $\widetilde{y}_s$ as the input to the depth completion module. The generator $G_d$ maps $\{x_d, x_i, \widetilde{y}_s\}$ to the fake dense depth image $\widetilde{y}_d$, and the generator $F_d$ remaps $\{\widetilde{y}_d, x_i, \widetilde{y}_s\}$ to the reconstructed sparse depth $\hat{x}_d$. Similarly, $F_d$ maps $\{y_d, x_i, \widetilde{y}_s\}$ to fake the sparse depth $\widetilde{x}_d$, and then $G_d$ maps $\{\widetilde{x}_d, x_i, \widetilde{y}_s\}$ to the reconstructed dense depth image $\hat{y}_d$. Similar to the semantic segmentation module, the discriminators $D_{X_{\scriptsize 2}}$ and $D_{Y_{\scriptsize 2}}$ in the depth completion module are used to distinguish real samples from $x_d$ and $y_d$ or generated samples $\widetilde{x}_d$ and $\widetilde{y}_d$. We apply the adversarial loss, cycle consistency loss, and reconstruction loss to improve the results of depth completion. We will not go into details here because they are similar to the corresponding losses in semantic segmentation. We will introduce several loss functions specifically for the depth completion task, which are different from the transfer branch of semantic labels. Letting $\mathcal{L}_{dep} \triangleq \mathcal{L}_{dep}(G_d, F_d, D_{X_{\scriptsize 2}}, D_{Y_{\scriptsize 2}})$, the overall loss function of the depth completion branch is as follows:
\begin{equation}\label{depth}
\begin{aligned}
\mathcal{L}_{dep} = & \mathcal{L}_{GAN}(G_d, D_{Y_{\scriptsize 2}}) + \mathcal{L}_{GAN}(F_d, D_{X_{\scriptsize 2}}) \\ & + \lambda_3 \mathcal{L}_{cyc}(G_d, F_d)  + \lambda_4 \mathcal{L}_{rec}(G_d, F_d)  \\ & + \lambda_5 \mathcal{L}_d(\widetilde{y}_d, x_d) + \lambda_6 \mathcal{L}_{smooth}(\widetilde{y}_d, \widetilde{y}_s),
\end{aligned}
\end{equation}
where $\lambda_3$, $\lambda_4$, $\lambda_5$, $\lambda_6$ are hyperparameter that control the weight of each loss. $\mathcal{L}_{GAN}(G_d, D_{Y_{\scriptsize 2}})$ and $\mathcal{L}_{GAN}(F_d, D_{X_{\scriptsize 2}})$ are adversarial losses \cite{goodfellow2014generative}. $\mathcal{L}_{cyc}(G_d, F_d)$ and $\mathcal{L}_{rec}(G_d, F_d)$ are the cycle consistency loss \cite{zhu2017unpaired} and the structural similarity reconstruction loss \cite{wang2004image} respectively, which are similar to the semantic segmentation branch. $\mathcal{L}_d(\widetilde{y}_d, x_d)$ stands for the depth loss \cite{ma2019self}. $\mathcal{L}_{smooth}(\widetilde{y}_d, \widetilde{y}_s)$ refers to the developed semantic-guided smoothness loss.

\textbf{Depth loss.} We take the sparse depth $x_d$ as the constraint of the generated dense depth image $\widetilde{y}_d$. We set the depth loss \cite{ma2019self} to penalize the difference between $x_d$ and $\widetilde{y}_d$, which will make the depth completion result more accurate. Leading to the depth loss:
\begin{equation}
	\mathcal{L}_d(\widetilde{y}_d, x_d) = \left\|1_{\{x_{\scriptsize d} >0\}} \cdot (\widetilde{y}_d - x_d)\right\|_2^2,
\end{equation}
where $1_{\{x_{\scriptsize d} >0\}}$ limits the effective point of the depth loss to the pixels with depth values in the sparse depth image.

\textbf{Semantic-Guided Smoothness loss.} In order to improve depth completion results, we develop the semantic-guided smoothness loss to constrain the smoothness of the generated dense depth images, which is inspired by \cite{chen2019towards} and \cite{zhao2020masked}. \cite{zhao2020masked} only considers the RGB image information to improve depth estimation results, while \cite{chen2019towards} uses semantic information to improve the depth estimation result only through simple operations such as pixel shifting and maximizing pixel value.  We develop the second-order differential \cite{zhao2020masked} for improving the smoothness at the semantic level, which reduce the impact of optical changes. The semantic-guided smoothness loss is:
\begin{equation}
	\mathcal{L}_{smooth}(\widetilde{y}_d, \widetilde{y}_s) = \sum_{p_t} | \nabla(\nabla \widetilde{y}_d(p_t)) | (e^{-|\nabla \widetilde{y}_s(p_t)|})^T,
\end{equation}
where $\widetilde{y}_d(p_t)$ and $\widetilde{y}_s(p_t)$ represent all pixels in $\widetilde{y}_d$ and $\widetilde{y}_s$, respectively. As well, $\nabla$ is a vector differential operator, and $T$ is the transpose operation.

\section{Experiments}\label{sec.4}

To evaluate our multi-task architecture, we carry out two experiments: semantic segmentation and depth completion. For both tasks, we use different inputs and similar generators. In this section, we will detail the experimental implementation, parameter settings, evaluation methods and evaluation results.

\subsection{Datasets}
\textbf{Cityscapes.} For the semantic segmentation task, we conduct a series of experiments on the Cityscapes dataset \cite{cordts2016cityscapes}. This dataset contains 2975 training images and 500 validation images, which have pixel-level annotations. Considering the training time and running cost, we adjust the data size from 2048$\times$1024 to 128$\times$128, which makes our experiment fair compared with CycleGAN. When generating a semantic label for depth completion, we adjust the image size to 512$\times$256 in order to match the resolution of the input image for depth completion. It is worth noting that although the images in the Cityscapes dataset are paired, we ignore the pairing information during training, i.e., our method is regarded as semi-supervised.

\textbf{KITTI.} For the depth completion task, we use the KITTI benchmark of depth completion for training and testing \cite{uhrig2017sparsity}. This benchmark of depth completion contains sparse depth images collected by LiDAR. It contains 85,898 training samples, 1,000 verification samples, and 1,000 testing samples. Each sparse depth has a corresponding RGB image. In order to reduce the training cost and time, we change the image size from 1216$\times$352 to 512$\times$256 during training, and we only use 5,815 samples as the training set. Since  discriminators of GANs are used to distinguish ground-truth and generated images, we need to input several dense depth images into the discriminator. In this paper, the dense depth images come from the KITTI raw dataset \cite{geiger2013vision}. During the evaluation, we use the scale consistency factor to align the scale of the completion result to the ground truth \cite{zhao2020masked}.

\newcommand{\tabincell}[2]{\begin{tabular}{@{}#1@{}}#2\end{tabular}}

\begin{table*}\normalsize
	\centering
	\begin{threeparttable}
		\caption{Semantic segmentation scores on Cityscapes}
		\begin{tabular}{c | c | c | c | c | c | c | c}
			\toprule
			\multirow{2}{*}{Year} & \multirow{2}{*}{Method} & \multicolumn{3}{c|}{Image$\rightarrow$Label} & \multicolumn{3}{c}{Label$\rightarrow$Image} \\
			\cline{3-8}
			& & Per-pixel acc. & Per-class acc. & Class IoU & Per-pixel acc. & Per-class acc. & Class IoU \\
			\midrule
			2016, NIPS & CoGAN \cite{liu2016coupled} & 0.45 & 0.11 & 0.08 & 0.40 & 0.10 & 0.06 \\
			2017, CVPR & SimGAN \cite{shrivastava2017learning} & 0.47 & 0.11 & 0.07 & 0.20 & 0.10 & 0.04 \\
			2017, ICCV & DualGAN \cite{yi2017dualgan} & 0.49 & 0.11 & 0.08 & 0.46 & 0.11 & 0.07 \\
			2017, ICCV & CycleGAN \cite{zhu2017unpaired} & 0.58 & 0.22 & 0.16 & 0.52 & 0.17 & 0.11 \\
			2019, CVPR & GcGAN-rot \cite{fu2019geometry} & 0.574 & 0.234 & 0.170 & 0.551 & 0.197 & 0.129 \\
			2019, CVPR & GcGAN-vf \cite{fu2019geometry} & 0.576 & 0.232 & 0.171 & 0.548 & 0.196 & 0.127 \\
			& Our &  \textbf{0.623} & \textbf{0.258} & \textbf{0.176} & \textbf{0.608}  & \textbf{0.243}  &  \textbf{0.159} \\
			\midrule
			\multicolumn{8}{c}{Ablation Study} \\
			\midrule
			2017, ICCV & CycleGAN \cite{zhu2017unpaired} & 0.58 & 0.22 & 0.16 & 0.52 & 0.17 & 0.11 \\
			& +MSSP &0.542 &0.242 &0.158 &0.496 &0.203 &0.138 \\
			& +$\mathcal{L}_{rec}$ &0.494 &0.193 &0.149 &0.469 &0.176 & 0.120\\
			& Total &\textbf{0.623} & \textbf{0.258} & \textbf{0.176}&\textbf{0.608} &\textbf{0.243} &\textbf{0.159} \\
			\bottomrule
		\end{tabular}
		\label{table1}
	\end{threeparttable}
\end{table*}

\begin{figure*}[htbp]
	\centering
	\includegraphics[width=18cm,height=7cm]{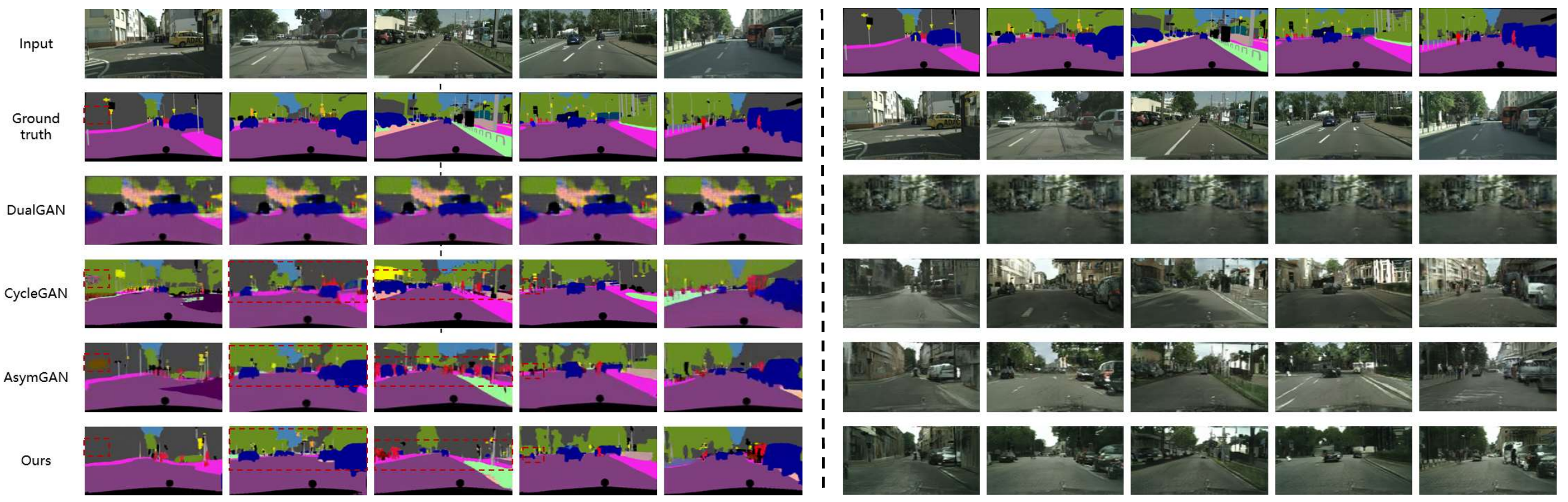}
	\caption{Qualitative comparison of generation quality on Cityscapes (Semantic label $\rightleftharpoons$ RGB image) between DualGAN \cite{yi2017dualgan}, CycleGAN \cite{zhu2017unpaired}, AsymGAN \cite{li2019asymmetric} and our method.}
	\label{Figseg1}
\end{figure*}

\begin{figure*}[htbp]
	\centering
	\includegraphics[width=18cm,height=7cm]{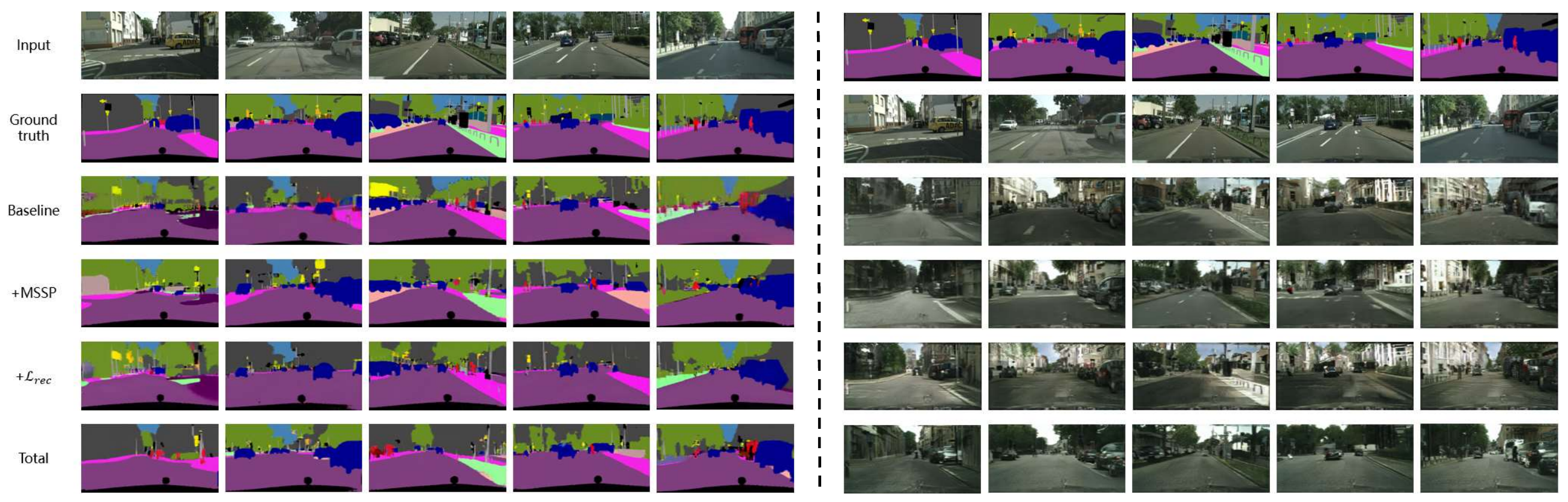}
	\caption{Ablation study. The influence of multi-scale spatial pooling blocks and the reconstruction loss on our method, based on CycleGAN.}
	\label{Figseg3}
\end{figure*}

\begin{figure*}[htbp]
	\centering
	\includegraphics[width=16cm,height=8.5cm]{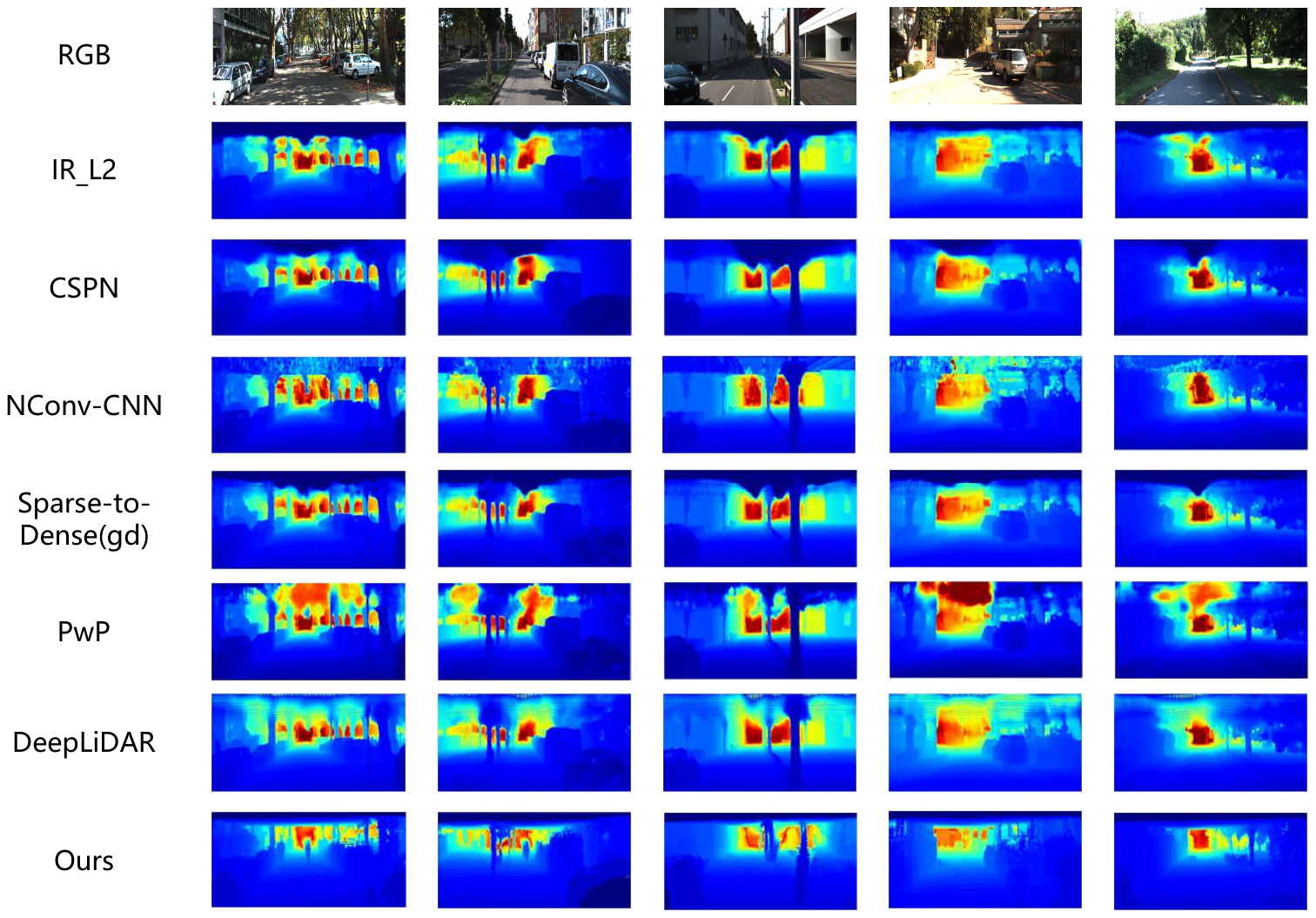}
	\caption{Qualitative comparison of depth completion quality on KITTI depth completion benchmark between IR\_L2 \cite{lu2020depth}, CSPN \cite{cheng2018depth}, NConv-CNN \cite{eldesokey2019confidence}, Sparse-to-Dense(gd) \cite{ma2019self}, PwP \cite{xu2019depth}, DeepLiDAR \cite{qiu2019deeplidar} and our method.}
	\label{Figdep1}
\end{figure*}

\begin{figure*}[htbp]
	\centering
	\includegraphics[width=16cm,height=8cm]{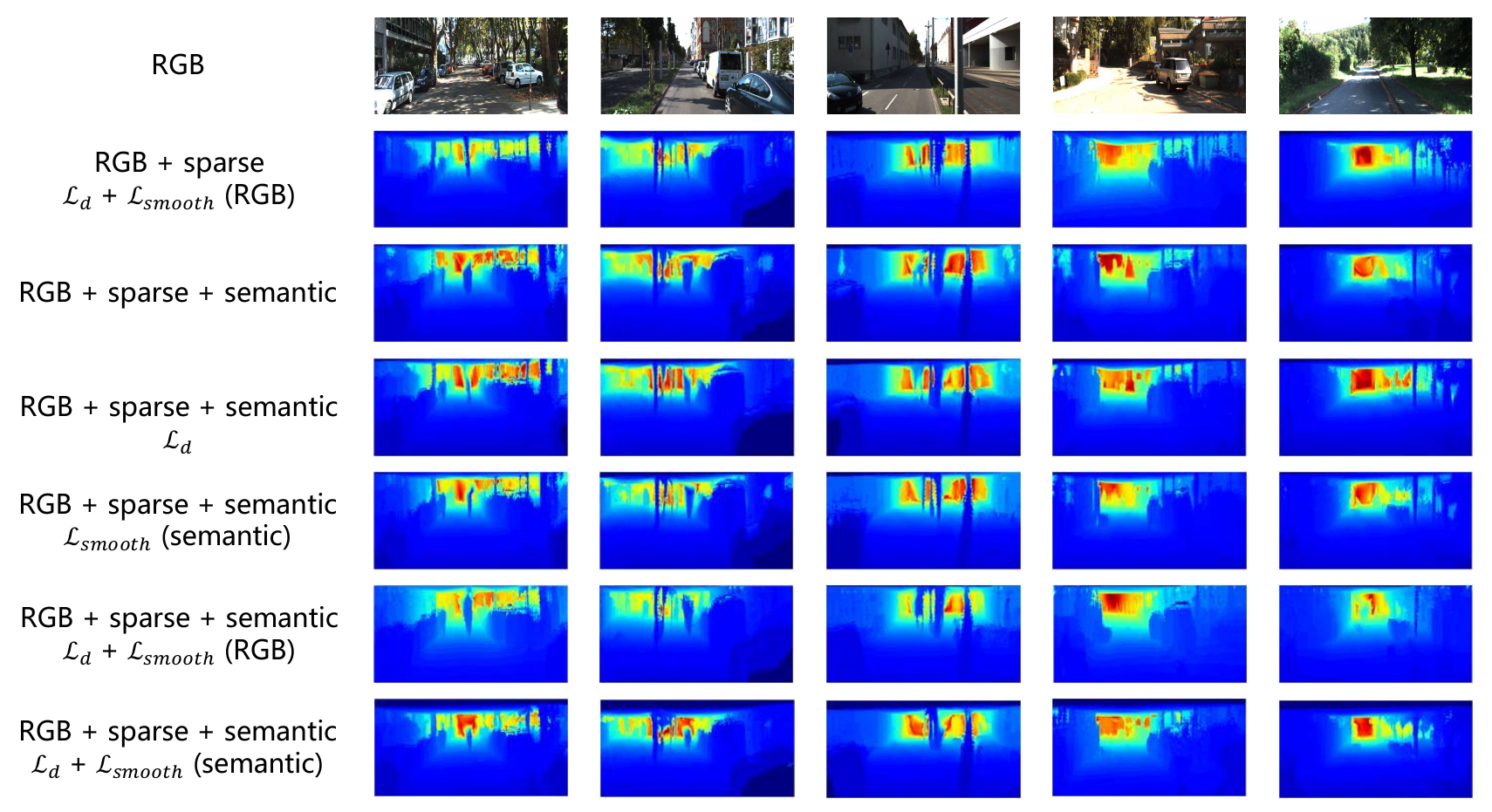}
	\caption{Ablation study. The influence of semantic information, depth loss and semantic-guided smoothness loss on our method.}
	\label{Figdep2}
\end{figure*}

\begin{table*}
	\centering
	\begin{threeparttable}
		\caption{Performance of depth completion on KITTI depth completion benchmark }
		\begin{tabular}{c | c | c | c | c | c | c }
			\toprule
			Year & Method & Input & RMSE [mm] $\downarrow$\tnote{1} & MAE [mm] $\downarrow$ & iRMSE [1/km] $\downarrow$ & iMAE [1/km] $\downarrow$ \\
			\midrule
			2020, CVPR & IR\_L2 \cite{lu2020depth} & depth-only & 901.43 & 292.36 & 4.92 & 1.35 \\
			2018, ECCV & CSPN \cite{cheng2018depth} & multiple-input & 1019.64 & 279.46 & 2.93 & 1.15\\
			2019, IEEE TPAMI & NConv-CNN \cite{eldesokey2019confidence} & multiple-input & 829.98 & 233.26 & 2.60 & \textbf{1.03}\\
			2019, ICRA & Sparse-to-Dense(gd) \cite{ma2019self} & multiple-input & 814.73 & 249.95 & 2.80 & 1.21\\
			2019, ICCV & PwP \cite{xu2019depth} & multiple-input & 777.05 & 235.17 & 2.42 & 1.13\\
			2019, CVPR & DeepLiDAR \cite{qiu2019deeplidar} & multiple-input & 758.38 & \textbf{226.50} & 2.56 & 1.15\\
			& Our & multiple-input & \textbf{746.96} & 267.71 & \textbf{2.24} & 1.10 \\
			\midrule
			\multicolumn{7}{c}{Ablation Study} \\
			\midrule
			Input\tnote{2} & \multicolumn{2}{c|}{Loss\tnote{3}} & RMSE [mm] $\downarrow$ & MAE [mm] $\downarrow$ & iRMSE [1/km] $\downarrow$ & iMAE [1/km] $\downarrow$ \\
			\midrule
			RGB + sparse & \multicolumn{2}{c|}{$\mathcal{L}_d$ + $\mathcal{L}_{smooth}$ (RGB)} &752.94 &299.82 &2.77 &1.24 \\
			RGB + sparse + semantic &\multicolumn{2}{c|}{} &869.01 &285.58 &3.00 &1.39 \\
			RGB + sparse + semantic & \multicolumn{2}{c|}{$\mathcal{L}_d$} &818.48 &288.71 &2.50 &1.45 \\
			RGB + sparse + semantic & \multicolumn{2}{c|}{$\mathcal{L}_{smooth}$ (semantic)}&784.87 &284.62 &2.50 &1.16 \\
			RGB + sparse + semantic & \multicolumn{2}{c|}{$\mathcal{L}_d$ + $\mathcal{L}_{smooth}$ (RGB)} &767.99 &292.62 &2.73 &1.61 \\
			RGB + sparse + semantic & \multicolumn{2}{c|}{$\mathcal{L}_d$ + $\mathcal{L}_{smooth}$ (semantic)}&\textbf{746.96} & \textbf{267.71} & \textbf{2.24} & \textbf{1.10} \\
			\bottomrule
		\end{tabular}
		\label{table2}
		\begin{tablenotes}
			\footnotesize
			\item[1] $\downarrow$ means smaller is better.
			\item[2] This column shows different inputs, ``RGB" represents  RGB images, ``sparse" represents sparse depth images, and ``semantic" represents the generated semantic images.
			\item[3] This column shows the different loss functions. ``$\mathcal{L}_d$" represents the depth loss, ``$\mathcal{L}_{smooth}$ (RGB)" represents the RGB-guided smoothness loss, and ``$\mathcal{L}_{smooth}$ (semantic)" represents the semantic-guided smoothness loss.
		\end{tablenotes}
	\end{threeparttable}
\end{table*}

\subsection{Semantic segmentation}
\textbf{Implementation.} For the semantic preception task, our generators $G_s$ and $F_s$ are adjusted based on CycleGAN \cite{zhu2017unpaired}. The generators $G_s$ and $F_s$ contain two stride-2 convolutions, six residual blocks, two $\frac{1}{2}$-strided convolutions \cite{zhu2017unpaired} and a multi-scale spatial pooling block \cite{tang2019multi}. Among them, the multi-scale spatial pooling block takes the concatenation of the RGB image and the convolution-residual-convolution output as input, and it contains a set of different kernel sizes for pooling to capture multi-scale features with different receptive fields. In this paper, the sizes of the different kernel we adopt are $(1,1)$, $(4,4)$ and $(9,9)$ \cite{tang2019multi}. For the discriminators $D_{X_{\scriptsize 1}}$ and $D_{Y_{\scriptsize 1}}$, we use $70 \times 70$ PatchGAN \cite{zhu2017unpaired}, which is the same as CycleGAN. For the semantic preception network, we set $\lambda_1 = 10$, $\lambda_2 = 2$. The hyperparameters of the semantic segmentation module are set according to experience and appropriately drawn on CycleGAN's parameter settings. During the training process, we set the learning rate to keep it at 0.0002 for the first 100 epochs, and linearly decay to 0 in the next 100 epochs.

\textbf{Qualitative evaluation.} The qualitative results are shown in Figure \ref{Figseg1}. We compare the results with DualGAN \cite{yi2017dualgan}, CycleGAN \cite{zhu2017unpaired} and AsymGAN \cite{li2019asymmetric}. DualGAN suffers from mode collapse, which produces an identical output regardless of the input image. CycleGAN may confuse vegetation and building labels and has a poor understanding of object details. AsymGAN performs well in details, but it sometimes generates objects that do not exist in the ground truth, as shown in Figure \ref{Figseg1}. It is worth noting that AsymGAN introduces an auxiliary variable by adding an encoder and a discriminator to learn the details, which greatly increases the complexity of the framework. For our method, we achieve comparable results to AsymGAN only by adding multi-scale spatial pooling blocks and the reconstruction loss on the basis of CycleGAN \cite{zhu2017unpaired}, which is less complex than AsymGAN because there is no need to adjust the parameters of the additional encoder and discriminator. Specifically, since the multi-scale spatial pooling block captures image features of different scales, and the structural similarity reconstruction loss restricts generated results by making the reconstructed image as similar to the original image as possible. Hence, we handle the problem that vegetation and buildings are sometimes turned over. Moreover, it is also competitive for the generated details, like pedestrians.


\textbf{Quantitative evaluation.} It is worth noting that since our semantic segmentation task is not strictly a semantic segmentation, the results need to be processed during the quantitative evaluation process, which is the same as \cite{zhu2017unpaired}, \cite{fu2019geometry}. Specifically, for semantic label$\rightarrow$image, we believe that input a high-quality generated image into a scene parser will produce good semantic segmentation results. Thus, we use the pretrained FCN-8s \cite{long2015fully} provided by pix2pix \cite{isola2017image} to predict the semantic labels for generated RGB images. In the quantitative evaluation process, the predicted labels are resized to be the same as the ground truth labels. Finally, the standard metrics of the predicted labels and ground truth labels are calculated, including pixel accuracy, class accuracy, and mean IoU \cite{long2015fully}. For image$\rightarrow$semantic label, since generated semantic images are in RGB format, which are not strictly equal to the RGB values corresponding to the ground truth. We should first align the RGB values of generated semantic images to the standard RGB values, and then map them to the class-level labels \cite{zhu2017unpaired}. Actually, we consider 19 category labels and 1 ignored label provided by Cityscapes, where each label corresponds to a standard color value. We compute the distance between each pixel of the generated semantic image and the standard RGB value, and we align the generated semantic image with the smallest distance label \cite{zhu2017unpaired}. Then we calculate the standard metrics for the generated semantic images and ground truth labels, which also include pixel accuracy, class accuracy, and mean IoU for evaluation.

The quantitative evaluation results are shown in Table \ref{table1}. Our method scores higher than the other methods for the quantitative evaluation of the image$\rightleftharpoons$semantic label on the Cityscapes dataset, including CoGAN \cite{liu2016coupled}, SimGAN \cite{shrivastava2017learning}, DualGAN \cite{yi2017dualgan}, CycleGAN \cite{zhu2017unpaired}, and GcGAN \cite{fu2019geometry}. For image$\rightarrow$semantic label, the results of our method are respectively improved by $4.7\%\sim17.3\%$, $2.6\%\sim14.8\%$ and $0.5\%\sim10.6\%$ compared with other methods for pixel accuracy, class accuracy, and mean IoU. For semantic label$\rightarrow$image, our method respectively improved by $5.7\%\sim40.8\%$, $4.6\%\sim14.3\%$ and $3\%\sim11.9\%$ compared with other methods for the above metrics. Experimental results show that our proposed method is competitive.

\textbf{Ablation study.} We study different ablations to analyze the effectiveness of multi-scale spatial pooling blocks and the reconstruction loss. Our qualitative results are shown in Figure \ref{Figseg3}, and quantitative results are given in Table \ref{table1}. We use the quantitative evaluation results of CycleGAN \cite{zhu2017unpaired} as a baseline. In Figure \ref{Figseg3} and Table \ref{table1}, ``$+$MSSP" denotes adding a multi-scale spatial pooling block based on CycleGAN, ``$+\mathcal{L}_{rec}$" refers to the addition of structural similarity reconstruction loss based on CycleGAN, and ``Total" is that both are added to the module at the same time. Figure \ref{Figseg3} shows that the multi-scale spatial pooling block is effective for capturing details. In addition, when multi-scale spatial pooling blocks and the reconstruction loss are added at the same time, the problem of confusion between buildings and vegetation is dealt with. Table \ref{table1} shows that although the results are not improved when multi-scale pooling blocks and the reconstruction loss are added separately, but when both are added to the framework at the same time, the results are significantly improved.

\subsection{Depth completion}
\textbf{Implementation.} For the depth completion task, we use sparse depth images, RGB images and generated semantic images as input, which are fed into the generator $G_d$ to obtain dense depth. Similarly, we use the generated dense depth, RGB images and generated semantic images as input to the generator $F_d$ to generate the corresponding sparse depth. For the depth completion branch, we only adjust the input method of the network. Specifically, we input the concatenation of three inputs and then pass through $G_d$ or $F_d$, which contain two stride-2 convolutions, nine residual blocks and two $\frac{1}{2}$-strided convolutions. For the discriminators $D_{X_{\scriptsize 2}}$ and $D_{Y_{\scriptsize 2}}$, we also use $70 \times 70$ PatchGAN \cite{zhu2017unpaired}. For the depth completion network, we set $\lambda_3 = 10$, $\lambda_4 = 1$, $\lambda_5 = 0.5$, $\lambda_6 = 0.5$. As in the semantic segmentation module, the learning rate is 0.0002 in the first 100 epochs, and then linearly decays to 0 for the next 100 epochs.

\textbf{Qualitative evaluation.} Qualitative comparisons are shown in Figure \ref{Figdep1}. We compare the results with IR\_L2 \cite{lu2020depth}, CSPN \cite{cheng2018depth}, NConv-CNN \cite{eldesokey2019confidence}, Sparse-to-Dense(gd) \cite{ma2019self}, PwP \cite{xu2019depth} and DeepLiDAR \cite{qiu2019deeplidar}. We find that IR\_L2 \cite{lu2020depth}, NConv-CNN \cite{eldesokey2019confidence}, and PwP \cite{xu2019depth} have poor effects on the upper sparse point completion, which show very noisy details. CSPN \cite{cheng2018depth} and Sparse-to-Dense (gd) \cite{ma2019self} do not completely complement the sparse points in a few upper regions. DeepLiDAR \cite{qiu2019deeplidar} produces more accurate depth with better details, like the completion of small objects, but the completion of the upper layer points will produce noisy results. In contrast, our method has better results for the depth complementation of the distant and upper layers, and the smoothness is better. This is due to our full use of semantic information, which is not sensitive to obvious changes in light in the distance. 

\textbf{Quantitative evaluation.} In this paper, we use the official metrics to quantitatively evaluate for the KITTI bechmark of depth completion. The four metrics include: root mean squared error (RMSE, mm), mean absolute error (MAE, mm), root mean squared error of the inverse depth (iRMSE, 1/km) and mean absolute error of the inverse depth (iMAE, 1/km). Among them, RMSE is the most important evaluation metric \cite{qiu2019deeplidar}. These metrics are formulated as:
\begin{itemize}
	\item \textbf{RMSE} = $\sqrt{\frac{1}{n}\sum_{i=1}^{n}(d_i-d_i^*)^2}$,
	\item \textbf{MAE} = $\frac{1}{n}\sum_{i=1}^{n}|d_i-d_i^*|$,
	\item \textbf{iRMSE} = $\sqrt{\frac{1}{n}\sum_{i=1}^{n}(\frac{1}{d_i}-\frac{1}{d_i^*})^2}$,
	\item \textbf{iMAE} = $\frac{1}{n}\sum_{i=1}^{n}|\frac{1}{d_i}-\frac{1}{d_i^*}|$,
\end{itemize}
where $d_i$ and $d_i^*$ stand for the predicted depth of pixel $i$ and corresponding ground truth. $n$ denotes the total number of pixels.

The quantitative results are reported in Table \ref{table2}. We compare our method with IR\_L2 \cite{lu2020depth}, CSPN \cite{cheng2018depth}, NConv-CNN \cite{eldesokey2019confidence}, Sparse-to-Dense(gd) \cite{ma2019self}, PwP \cite{xu2019depth} and DeepLiDAR \cite{qiu2019deeplidar}. For RMSE and iRMSE, our method is better than the other methods. The iMAE of our method is better than the other methods except NConv-CNN \cite{eldesokey2019confidence}. The MAE index is slightly inferior, which may because the ground truth input by our method comes from the KITTI raw dataset, which does not match the scale of the KITTI benchmark for depth completion.

\textbf{Ablation study.} In order to understand the impact of semantic information and various loss functions on the final performance, we disable the semantic input and each loss function and show how the result changes. The qualitative results are shown in Figure \ref{Figdep2}, and quantitative results in Table \ref{table2}. The first row shows that no semantic information is introduced. Compared with the introduction of semantic information, which is shown in the last row, it can be seen that semantic information is helpful for improving the four indicators. In order to verify the effectiveness of depth loss, we conduct two sets of comparative experiments. The results of the second and third rows are used to verify that only the depth loss is added when the RGB images, sparse depth images, and semantic labels are used as input. The results of the fourth and sixth rows are used to verify the effect of depth loss on the results when the above three are used as inputs and the semantic-guided smoothness loss is added. Both sets of comparative experiments prove that depth loss can improve experimental results. We also conduct two sets of controlled experiments to verify the effectiveness of the semantic-guided smoothness loss. By comparing the results of the third row and the sixth row, we see that the accuracy is improved when the semantic-guided smoothness loss is added. As well, by comparing the results of the fifth and sixth rows, we see that adding the semantic-guided smoothness loss is more significant for the result than adding the RGB-guided smoothness loss. The above ablation studies confirm that our improvements to depth complement are effective.

\section{Conclusion}\label{sec.5}

In this paper, we propose multi-task generative adversarial networks (Multi-task GANs), including a semantic segmentation module and a depth completion module. For semantic segmentation task, we introduce multi-scale spatial pooling blocks to extract different scale features of images, which effectively tackle the problem of poor details generated by image translation. In addition, we add the structural similarity reconstruction loss to further improve semantic segmentation results from the perspectives of luminance, contrast and structure. Moreover, we take the concatenation of the generated semantic image, the sparse depth, and the RGB image for the input of depth completion module, and we introduce the semantic-guided smoothness loss to improve the result of depth completion. Our experiments show that the introduction of semantic information effectively improves the accuracy of depth completion. In the future, we will share the network layers of semantic segmentation and depth completion, and we will use the generated images to promote each other.



  \label{}



\bibliographystyle{IEEEtran}
\bibliography{Ref}




\end{document}